%% file: main.tex
\documentclass[10pt,twocolumn,letterpaper]{article}

\usepackage[pagenumbers]{cvpr} 
\usepackage[ruled,vlined]{algorithm2e} 
\usepackage{amsmath}
\input{preamble}

\definecolor{cvprblue}{rgb}{0.21,0.49,0.74}
\usepackage[pagebackref,breaklinks,colorlinks,allcolors=cvprblue]{hyperref}

\title{ABE--CLIP: Training-Free Attribute Binding Enhancement\\
for Compositional Image--Text Matching}

\author{
Qi Zhang\textsuperscript{1},
Yuxu Chen\textsuperscript{2},
Lei Deng\textsuperscript{2},
Lili Shen\textsuperscript{1} \\
\textsuperscript{1}School of Mathematics, Sichuan University \\
\textsuperscript{2}Independent Researcher
}

\begin{document}
\maketitle
\input{sec/0_abstract}    
\input{sec/1_intro}
\input{sec/2_formatting}

\input{sec/3_finalcopy}
{
    \small
    \bibliographystyle{ieeenat_fullname}
    \bibliography{main}
}


\end{document}

%% file: preamble.tex
\usepackage[dvipsnames,table]{xcolor}

\usepackage{pifont}

\newcommand{\argtopk}{\operatorname*{arg\,\mathrm{top}-}\!K}
\usepackage{booktabs,siunitx}
\usepackage{caption}
\usepackage{subcaption}

\usepackage{tikz}

\usepackage{pgfplots}
\pgfplotsset{compat=1.18}
\usetikzlibrary{patterns}
\usepgfplotslibrary{groupplots} 

\usepackage[most]{tcolorbox}
\usepackage{graphicx,adjustbox}
\usepackage{array}

\definecolor{posc}{RGB}{78,170,94}   
\definecolor{negc}{RGB}{230,85,62}   
\definecolor{barframe}{RGB}{40,40,40}

\definecolor{myBlue}{RGB}{140,179,217}
\definecolor{myOrange}{RGB}{244,178,102}
\definecolor{myRed}{RGB}{224,122,106}

%% file: sec/0_abstract.tex
\begin{abstract}
Contrastive Language-Image Pretraining (CLIP) has achieved remarkable performance in various multimodal tasks. However, it still struggles with compositional image-text matching, particularly in accurately associating objects with their corresponding attributes, because its inherent global representation often overlooks the fine-grained semantic for attribute binding. 
Existing methods often require additional training or extensive hard-negative sampling, yet they frequently yield limited generalization ability to novel compositional concepts and struggle to fundamentally address the global representation's inherent drawbacks. 
In this paper, we propose ABE-CLIP, a novel training-free \textbf{A}ttribute \textbf{B}inding \textbf{E}nhancement method designed to strengthen the attribute-object binding capability of CLIP-like models. 
Specifically, we first employ the Semantic Refinement Mechanism to refine the token embedding for both object and attribute phrases within the text description, thereby mitigating attribute confusion and enhancing semantic precision. Subsequently, the Local Token-Patch Local Alignment is introduced to calculated the similarity scores of the refined  textual tokens to its most related patches. By intelligently aggregating the localized similarity scores, $\text{ABE-CLIP}$ computes the final image-text similarity.
Experiments on multiple datasets demonstrate that ABE-CLIP achieves significant improvements in attribute-object binding, even surpassing the methods that require extensive training.
\end{abstract}



%% file: sec/1_intro.tex
\section{Introduction}
\label{sec:intro}
Image--text matching~\citep{plummer2015flickr30k,lin2014microsoft} is a fundamental task in multimodal learning that aims to measure the semantic correspondence between visual content and textual descriptions. 
With recent advances in multimodal pre-training, contrastive learning---pioneered by CLIP~\citep{radford2021learning}---has emerges as one of the dominant paradigm. CLIP~\citep{radford2021learning} has demonstrated exceptional performance in multimodal tasks, such as image--text retrieval~\citep{pan2023prior,sun2024alpha,zhang2024long}, image captioning~\citep{mokady2021clipcap,li2024monkey,yao2024minicpm} and visual question answering~\citep{li2023blip,parelli2023clip,gemini2024unlocking,wang2025iaa}. Specifically, CLIP adopts a dual-tower architecture, where separate image encoder and text encoder separately extract visual and textual global feature vectors. The similarity between an image and a text is measured by the cosine similarity of their global feature vectors in a shared embedding space. This global vector representation enables high efficiency in retrieval tasks, making CLIP widely adopted in image-text retrieval scenarios.
\begin{figure}[t]
  \centering
  \includegraphics[width=\linewidth]{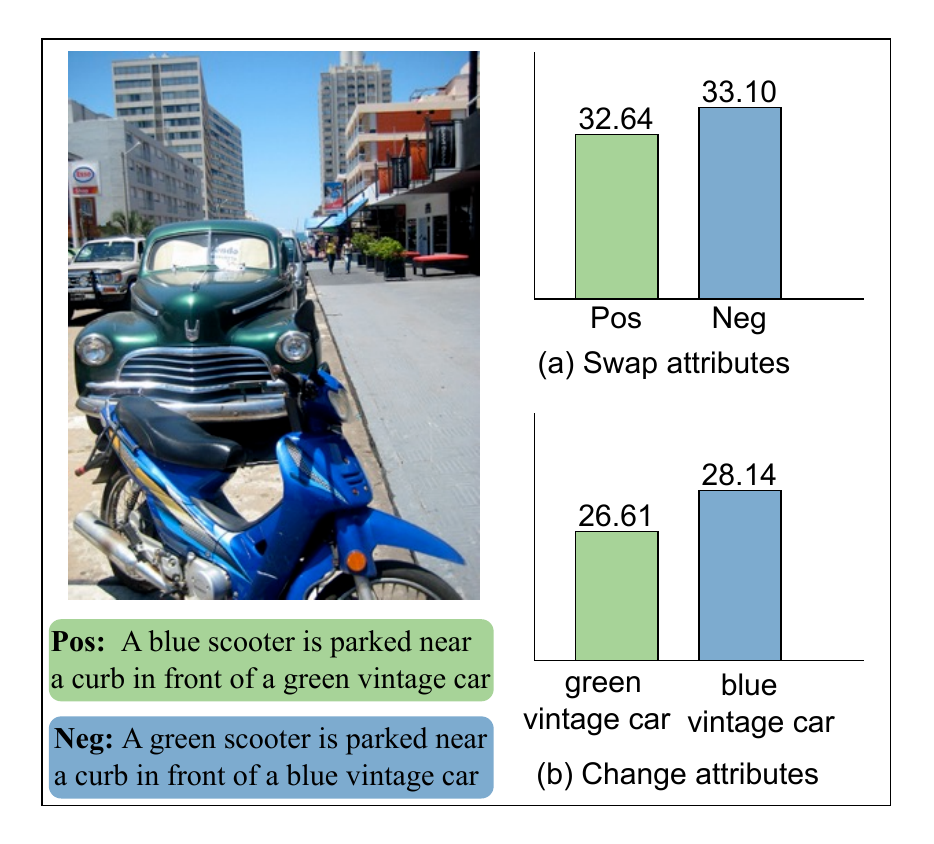}
  \caption{This illustration shows that CLIP fails to bind the attribute ``green'' to the corresponding object ``vintage car''. (a) The positive captions and the negative captions contain identical words while the attributes are swapped; (b) The CLIP similarity between the picture and the phrases ``green vintage car" and ``blue vintage car". }
  \label{fig:clip-fail-green-car}
\end{figure}

However, the global image-text embeddings learned by $\text{CLIP}$ often struggle with compositional semantic understanding, particularly when required to accurately bind attributes to objects in complex scenes with multiple entities. Evidence indicates that the global representation vectors of CLIP learns disentangled, bag-of-words-style representations~\citep{yuksekgonul2022and}, which fundamentally limits its ability to capture finer-grained compositional information. As illustrated in~\cref{fig:clip-fail-green-car}(a), CLIP struggles to distinguish the positive caption ``A blue scooter is parked near a curb in front of a green vintage car'' from the negative caption ``A green scooter is parked near a curb in front of a blue vintage car'' when attributes are swapped. This failure reveals that relying solely on the cosine similarity between CLIP’s global image and text embedding is insufficient for fine-grained semantic understanding in compositional image--text matching.

To overcome the aforementioned limitations in compositional understanding, various methods have been proposed to enhance the fine-grained matching capability of CLIP-like models. The predominant strategy among existing approaches is to leverage hard negative examples to force CLIP-like models to discern subtle textual or visual differences~\citep{kalantidis2020hard,zhang2024countercurate,awal2024vismin,yuksekgonul2022and,huang2024structure}. 
Although these methods have improved compositional understanding of CLIP-like models to some extent, they still yield limited generalization to unseen compositions. In practice, such pipelines often overfit to dataset-construction artifacts and remain brittle, frequently misjudging simple relations or swapping attributes across co-occurring objects under minor perturbations. 
Different from hard-negative sampling, OC-CLIP~\citep{assouel2025occlip} enhances the ability of CLIP-based architectures to learn attribute--object binding by incorporating an object-centric cross-modal binding module. However, its structured head increases architectural complexity and depends on accurate text-to-scene-graph parsing.

Alternatively, some compositional retrieval methods, like ComCLIP \cite{jiang2022comclip}, employ a strategy that decomposes captions into multiple phrases, retrieves each phrase independently, and ultimately aggregates the phrase-level similarity scores to mitigate CLIP’s limitations in addressing attribute binding and other compositional challenges. However, determining the optimal fine-grained level for phrase decomposition remains a non-trivial challenge. If attributes and their corresponding objects are split into separate phrases, compositional retrieval still fails to bind attributes to objects effectively. When an attribute and its associated object are decomposed into a single phrase, the CLIP model may still struggle to accurately match the attribute-object pair. For instance,  as demonstrated in~\cref{fig:clip-fail-green-car}(b),  the similarity score between the image and the phrase ``green vintage car"  is lower than that between the image and the phrase ``blue vintage car'', which is contradict to the ground truth. Therefore, simply decomposing captions is insufficient to effectively address the attribute binding problem.


In this work, we aim to enhance the attribute binding ability of CLIP-based models, a fundamental factor for compositional understanding of multiple objects in complex scenes. We propose a novel plug-in and training-free framework based on CLIP-like models, named \textbf{ABE-CLIP}, which overcome the limitations of the existing text decomposition-based methods.
Specifically, we propose a Local Token-Patch Alignment that computes the similarity between each text token and individual image patches, selects the patches with high similarities to each token, and thereby binds each text token to corresponding local patches. This design models the fine-grained semantic associations between textual units and image regions, while suppressing the interference from irrelevant patches, laying a solid foundation for accurate cross-modal alignment at the local level. 
To mitigate attribute confusion in the text modality, we incorporate a Semantic Refinement Mechanism that refines text embeddings specific to objects and attributes. This mechanism facilitates the disentanglement of object and attribute representations, yielding more discriminative vector embeddings for each text token.
Besides, we put forward a Binding Difference Scores to directly assess the effectiveness of representation refinement, and its output can serve as one component of the final image-text similarity score. 
Lastly, we leverage above components to design an overall image--text matching score. 
The widely experiments on multiple benchmarks shows that the proposed method achieves state-of-the-art performance compared to existing approaches.

Our contributions can be summarized as follows:
\begin{itemize}
  \item We propose \textbf{ABE-CLIP}, a training-free method that effectively enhances the attribute binding capability of CLIP-based models. As a plug-and-play framework, it can be seamlessly integrated into all CLIP-like models.

  \item  ABE-CLIP integrates representation refinement and local token-patch alignment, overcoming the limitations of previous text decomposition-based approaches.

  \item We conduct experiments on multiple image-text matching benchmarks, and the results demonstrate that our method achieves substantial improvements in attribute binding accuracy, outperforming other existing approaches.
\end{itemize}

%% file: sec/2_formatting.tex
\section{Preliminaries and Related Work}
\label{sec:related work}

\paragraph{Pretrained Contrastive Vision--Language Models.}
Vision--language models (VLMs) pretrained on large-scale image--text pairs have demonstrated impressive performance in multimodal domains~\citep{jia2021scaling,radford2021learning,li2022blip,mu2022slip,sun2024eva18b}. Following CLIP~\citep{radford2021learning}, contrastive pretraining has become a prevalent strategy that aligns image and text representations in a shared embedding space by pulling together matched image--text pairs and pushing apart mismatched pairs in a dual-encoder setup~\citep{jia2021scaling,zhai2022lit,zhai2023sigmoid,li2022blip,mu2022slip,sun2023eva}. 
However, such global alignment objectives often struggle to capture fine-grained semantic details, which is crucial for compositional image-text understanding-particularly in recognizing object attributes~\citep{thrush2022winoground,yuksekgonul2022and,zhao2022vl}. To address these limitations, a number of fine-grained approaches have been proposed~\citep{yao2021filip,zhong2022regionclip,patel2024tripletclip,xie2025fg}. For example,  FILIP~\citep{yao2021filip} proposes a fine-grained, cross-modal late-interaction objective that aligns text tokens with image patches by max-pooling token-level similarities. GLIP~\citep{li2022grounded} and RegionCLIP~\citep{zhong2022regionclip} pretrain on region–text pairs, thereby strengthening the model's capacity for fine-grained region-text correspondence. However, these methods often require additional training and are prone to bias induced by spurious correlations in the pretraining data.

\paragraph{Attribute Binding.} 
Attribute binding refers to a model's ability to correctly associate textual attributes with the  corresponding specific objects or regions within a visual scene, particularly in complex scenarios involving multiple objects and multiple attributes~\citep{thrush2022winoground,zhao2022vl,hsieh2023sugarcrepe,lewis2022does,yuksekgonul2022and}. In recent work, fine-tuning with hard negative examples has become a proven strategy to enhance compositionality~\citep{yuksekgonul2022and,zhang2024contrasting,huang2024structure}. Among them, NegCLIP~\citep{yuksekgonul2022and} synthesizes negative image--text pairs by sampling the nearest neighboring images and swapping caption words, and uses these negatives for fine-tuning. Additionally, 
Structure-CLIP~\citep{huang2024structure} integrates scene graph knowledge (SGK) to enhance structured representations, and leverages SGK to construct hard negative captions that better match the underlying intent. Moreover, LABCLIP~\citep{koishigarina2025clip} improves  attribute--object binding during cross-modal matching by applying a learned linear transformation to text embeddings, trained on hard negatives.  
Yet, these methods depend on hard-negative mining or enhancing structure representations, and still leave attribute--object binding implicit, leading to brittle improvements on unseen compositions.

\paragraph{Representation Refinement.}
Recent findings suggest that learned representations are decomposable and composable ~\citep{trager2023linear,berasi2025not,bhalla2024interpreting}. In particular, \citet{trager2023linear} shows that composite concepts can be approximated as a linear combination of embedding vectors associated with different factors, termed ideal words. \citet{berasi2025not} proposes a Geodesically Decomposable Embeddings framework that decomposes visual representations as a geometry-aware combination of optimal directions representing primitive concepts. These observations support representation refinement along semantically meaningful directions as a reasonable refinement strategy without retraining the backbone.
For example, within text-to-image generation, Magnet~\citep{zhuang2024magnet} modifies the object embeddings by pulling target attributes while pushing irrelevant ones, showing the ability to disentangle different attributes and generate anti-prior concepts. Inspired by Magnet, \citep{luo2025debiasing} refines object embeddings, and biased object terms are replaced with broader concepts during diffusion sampling to reduce bias. Nevertheless, leveraging representation refinement on textual embeddings for cross-modal retrieval remains under-explored.

\paragraph{CLIP} 
Contrastive Language–Image Pre-training (CLIP)~\cite{radford2021learning} has emerged as one of the dominant paradigms for VLM pretraining, significantly advancing image--text alignment through large-scale datasets~\cite{jia2021scaling,zhang2022contrastive,liu2023visual,dai2023instructblip,chen2022pali,cai2024vip,beyer2024paligemma}. It employs a dual-encoder architecture to project both modalities into a shared semantic space and utilizes a contrastive objective that pulls matched image--text pairs closer while pushing mismatched ones apart.

Specifically, given an input image $I$ and an input text $T$, the visual encoder $E_V(\cdot)$ first splits $I$ into $N$ non-overlapping patches, then outputs a global visual [CLS] token embedding $\mathbf{v}_{\text{cls}}$ (for whole-image representation) and $N$ local visual patch embeddings $\{\mathbf{v}_i\}_{i=1}^N$. Meanwhile, the text encoder $E_T(\cdot)$ processes $T$ into a sequence of up to $M$ textual tokens, generating a global textual [EOT] token embedding $\mathbf{t}_{\text{eot}}$ and $M$ local textual token embeddings $\{\mathbf{t}_j\}_{j=1}^M$, where each $\mathbf{t}_j$ corresponds to the $j$-th text token. CLIP models cross-modal interactions via global alignment between the entire image and its description. This approach lacks fine-grained correspondences between specific image regions and individual words or phrases. As a result, it struggles with accurate attribute--object binding, especially the phrase--region grounding.

\paragraph{FG-CLIP}  
To address this limitation, FG-CLIP~\cite{xie2025fg} extends CLIP with multi-granular visual--textual alignment through two key innovations. 
First, it introduces a regional contrastive loss to align region features with corresponding textual phrases. Second, it employs a hard negative mining strategy that rewrites descriptions to create semantically close negatives, sharpening fine-grained discrimination. This integrated approach empowers FG-CLIP's ability to match local text tokens with corresponding image regions, thereby enabling it to discern fine-grained visual--semantic details in complex compositional scenes, significantly boosting its performance on diverse downstream tasks.

\begin{figure*}[htbp]
  \centering
    \centering
    \includegraphics[width=\linewidth]{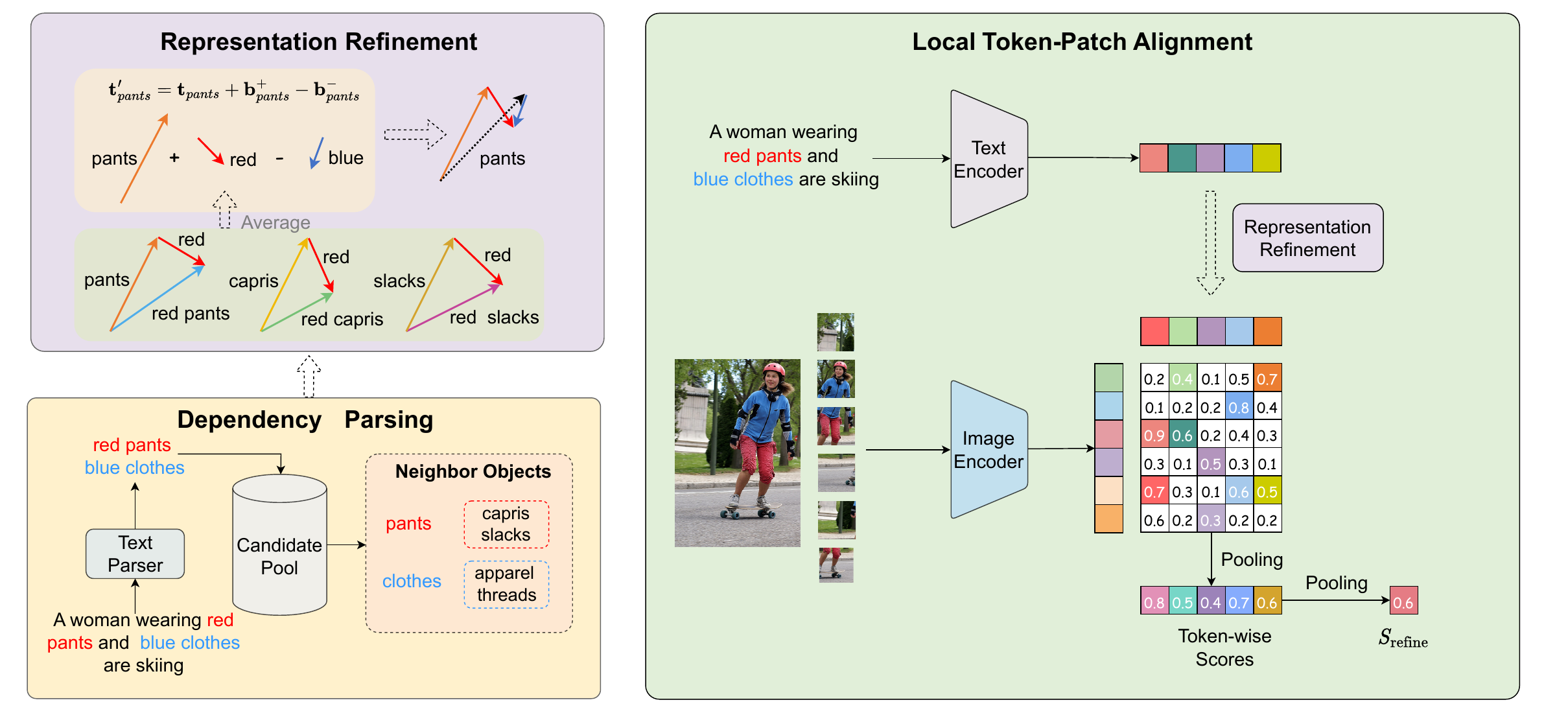}
  \caption{Overview of ABE-CLIP. Given a query caption, we first parse it to extract attribute--object phrases, then refine the attribute embedding and object embedding via Representation Refinement Mechanism. Subsequently, the Local Token-Patch Local Alignment is introduced to aggregate token--patch similarities.}
  \label{fig:pipeline}
\end{figure*}

\section{Proposed Method}
\label{sec:method}

The fundamental limitation of standard CLIP models in image-text matching ability stems from the inherent deficiency of its single global embedding~\cite{radford2021learning}. Although effective for high-level semantic alignment, the obligatory aggregation of image and text information results in a coarse, ``bag-of-words'' representation~\cite{yuksekgonul2022and}, where semantic components are linearly combine. Crucially, this aggregator nature inevitably blurs fine-grained details, making precise attribute--object binding alignment extremely difficult, severely restricting the model's compositional capability.

We introduce \textbf{ABE-CLIP}, a training-free framework designed to enhance the ability of CLIP-based models to correctly associate attributes with their corresponding objects.The proposed method adopts a local alignment strategy, which consists of the following steps:
\begin{itemize}
    \item Instead of relying on the similarity of coarse global vectors between images and texts, ABE-CLIP computes fine-grained token--patch similarities between the vector representations of individual text tokens and their corresponding image patches to accurately bind descriptive attributes to relevant visual objects, significantly boosting the model's performance in compositional understanding.
    \item To address the limitations imposed by entangled representations of text modal, we introduce a semantic refinement mechanism. This mechanism modifies the text embeddings to facilitate attribute-object disentanglement, thereby enabling a more precise matching between words and image patches.
    \item Finally, by integrating global similarity, refined local matching results and the binding difference, we  compute the final score to evaluate the image--text similarity.
\end{itemize}


\subsection{Local Token--Patch Alignment}
\label{sec:tpsm}
CLIP's global representation paradigm is limited in capturing compositional semantics, making it difficult to accurately associate specific attributes with the correct objects~\cite{zeng2021multi,lewis2022does}.
This is because global embeddings tend to conflate information from multiple objects and attributes within complex scenes, hindering the discrimination of specific attribute--object associations.
To alleviate this problem, we take a fine-grained token--patch alignment framework to model token-wise cross-modal interactions.

Specifically, given an image--text pair $(I,T)$, the CLIP model encodes the image $I$ into a sequence of visual patch embeddings $\mathbf{V}=\{\mathbf{v}_1, \dots, \mathbf{v}_{N}\}\in \mathbb{R}^{N\times d}$ and a visual global embedding $\mathbf{v}_{\text{cls}}$, and encodes the text $T$ into a sequence of textual token embeddings $\mathbf{W}=\{\mathbf{t}_1, \ldots, \mathbf{t}_{M}\}\in\mathbb{R}^{M\times d}$ and a textual global embedding $\mathbf{t}_{\text{eot}}$.
The semantic correspondence between textual tokens and visual patches is characterized by their cosine similarity:
\begin{equation}
\label{eq:t2p-sim}
S_{i,j}=\frac{\mathbf{t}_i^\top \mathbf{v}_j}{\|\mathbf{t}_i\|_2\,\|\mathbf{v}_j\|_2},\quad \mathbf{S}\in\mathbb{R}^{M\times N}.
\end{equation}

For each textual token $\mathbf{t}_i$, we select the top-$K$ most similar patches, forming an index set
\begin{equation}
\label{eq:t2p-topk}
\mathcal{P}_i \;=\; \argtopk \big\{\, S_{i,j} \,\big\}_{j=1}^{N},\qquad |\mathcal P_i|=K.
\end{equation}

Subsequently, the token-wise score for each textual token $\mathbf{t}_i$ is defined as the average cosine similarity over its corresponding top-$K$ patches:
\begin{equation}
\label{eq:t2p-phi}
\phi_i \;=\; \frac{1}{|\mathcal{P}_i|}\sum_{j\in\mathcal{P}_i} S_{i,j}.
\end{equation}

Then, the overall token--patch image--text matching score is computed by averaging the token-wise similarities across all text tokens:
\begin{equation}
\label{eq:t2p-psi}
\mathcal{S}_{\text{base}}(I,T) \;=\; \frac{1}{M}\sum_{i=1}^{M}\phi_i.
\end{equation}

Aggregating these local scores yields a finer-grained and more robust cross-modal alignment, thereby improving fine-grained discriminative power and interpretability in complex compositional scenes.


\subsection{Semantic Refinement Mechanism} 
\label{sec:semantic-refinement-mechanism}

\begin{figure}[t]
  \centering
  \newlength{\subfigH}\setlength{\subfigH}{0.16\textheight} 
  \newlength{\sep}\setlength{\sep}{0.02\columnwidth}       
  \begin{minipage}{0.98\columnwidth}
    \centering
    \begin{subfigure}[t]{0.30\columnwidth}
      \centering
      \includegraphics[width=\linewidth,height=\subfigH,keepaspectratio]{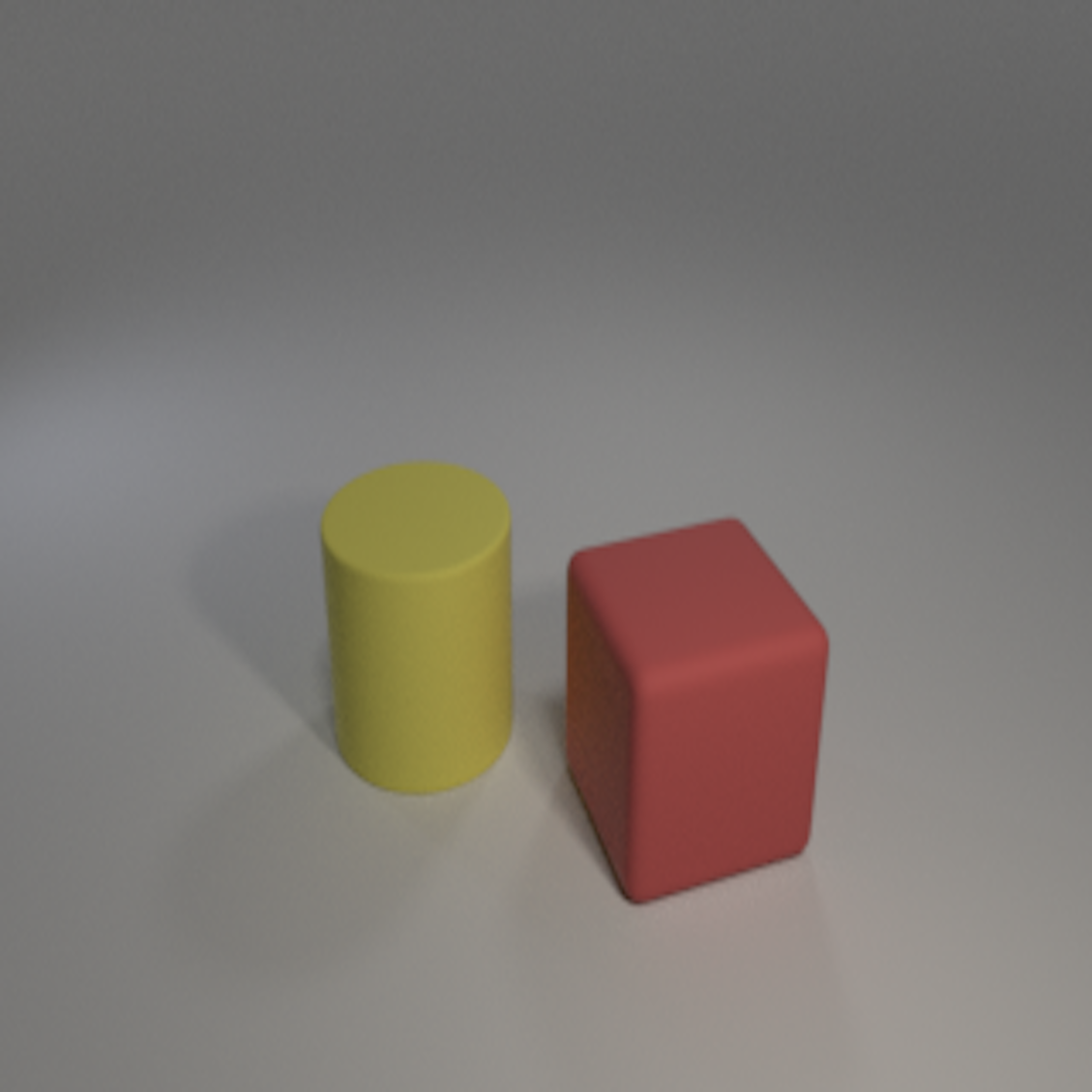}
      \caption{A yellow cylinder and a red cube.}\label{fig:img1}
    \end{subfigure}\hspace{\sep}
    \begin{subfigure}[t]{0.30\columnwidth}
      \centering
      \includegraphics[width=\linewidth,height=\subfigH,keepaspectratio]{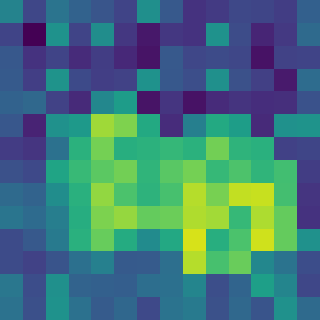}
      \caption{Pre-refinement}\label{fig:img2}
    \end{subfigure}\hspace{\sep}
    \begin{subfigure}[t]{0.30\columnwidth}
      \centering
      \includegraphics[width=\linewidth,height=\subfigH,keepaspectratio]{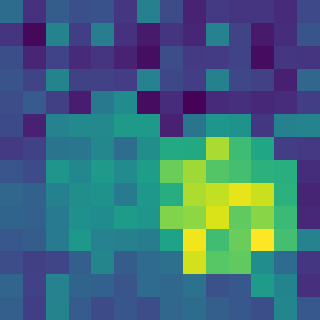}
      \caption{Post-refinement}\label{fig:img3}
    \end{subfigure}
  \end{minipage}

  \caption{\textbf{Representation refinement sharpens token--patch alignment for ``cube''.}
  (a) Input image ``A yellow cylinder and a red cube.'' (b,c) Similarity matrix visualizations before and after refinement between the attribute ``red'' and each image patch.}
  \label{fig:refinement-by-vector-reconstruction}
\end{figure}

However, relying solely on the local token--patch alignment is insufficient, as it often causes attribute tokens to be attracted by patches from the wrong object. For example, in ``A yellow cylinder and a red cube'', as shown in \cref{fig:refinement-by-vector-reconstruction} (a) and (b), using unrefined contextualized token embeddings can be problematic: the attribute token \emph{red} may be incorrectly bound to the object token \emph{cylinder} rather than to the \emph{cube}. Therefore, we need to modify the embeddings of the attribute and object tokens to decouple incorrect bindings and consolidate correct ones, so as to obtain more discriminative text representations. Inspired by the text-to-image generation model Magnet~\cite{zhuang2024magnet}, we instantiate binding vectors to explicitly steer object embeddings toward their target attributes while repelling them from irrelevant ones.

Specifically, we parse each caption $c$ with a natural language processing tool Stanza~\cite{qi2020stanza} to extract the attribute--object pairs $\mathcal{D}(c)=\{(a_m,k_m)\}_{m=1}^{M_c}$, where $M_c$ denotes the number of extracted pairs in caption $c$.
For each object $k$ in $\mathcal{D}(c)$ with the text embedding $\mathbf{t}_k$, we denote its target attribute $A_c^{+}(k)=a$ and the in-sentence unrelated attributes
$A_c^{-}(k)=\{a':(a',k')\in\mathcal{D}(c),\,k'\neq k\}$ respectively. We select the $P$ nearest neighbor objects of $k$ in the candidate pool $\mathcal{C}$ by cosine similarity as
$\mathcal{N}(k)=\{\mathrm{obj}_i\}_{i=1}^{P}$.

The positive and negative binding vectors of object $k$ are computed as follows:
\begin{equation}  
\mathbf{b}_k^{+} = \frac{1}{P} \sum_{i=1}^{P} \left( 
    \mathcal{F}(A_c^{+}(k), \mathrm{obj}_i) - 
    \mathcal{F}(\emptyset, \mathrm{obj}_i) 
\right)
\end{equation}
\begin{equation}  
\mathbf{b}_k^{-} = \frac{1}{P} \sum_{i=1}^{P} \left( 
   \mathcal{F}(A_c^{-}(k), \mathrm{obj}_i) - 
    \mathcal{F}(\emptyset, \mathrm{obj}_i) 
\right)
\end{equation}
where $\emptyset$ is a blank text, and \(\mathcal{F}(a,\mathrm{obj}_i)\) is the final-layer embedding of \(\mathrm{obj}_i\) extracted from the CLIP text encoder when encoding the phrase \((a,\mathrm{obj}_i)\). 
Unlike Magnet, which only employs a candidate pool of 614 object nouns, we adopt the more comprehensive and empirically-grounded conceptual vocabulary provided by SPLiCE~\cite{bhalla2024interpreting}, containing over 12K concepts. This broader conceptual basis can further improve the robustness and generalizability of complex open-world image-text alignment.  

We refine the text embedding of object $k$ via binding vectors:
 \begin{equation}
 \mathbf{t}'_k = \mathbf{t}_k + \mathbf{b}_k^{+} - \mathbf{b}_k^{-}
 \end{equation}

In addition, the attribute embedding $\mathbf{t}'_a$ in the initial text embeddings is modified by:
 \begin{equation}
 \mathbf{t}'_a = \mathbf{t}_a + \mathbf{t}_k
 \end{equation}

We incorporate the modified text embeddings into the token--patch alignment of Section~\ref{sec:tpsm}, yielding refined similarity scores, denoted $\mathcal{S}_{\text{refine}}(I,T)$.

\begin{algorithm}[h]
\caption{ABE-CLIP}
\label{alg:abe-clip}
\KwIn{Image $I$, Text $T$, Pretrained CLIP $(E_V, E_T)$, parameters $K$, $\omega$}
\KwOut{Final similarity score $\mathcal{S}_{\mathrm{final}}(I,T)$}
\vspace{0.5em}
\textbf{Step 1: Compute Original Similarities} 

$\{\mathbf{v}_i\}_{i=1}^N, \mathbf{v}_{\text{cls}} \leftarrow E_V(I)$\;
$\{\mathbf{t}_j\}_{j=1}^M, \mathbf{t}_{\text{eot}} \leftarrow E_T(T)$\;

Global similarity $\mathcal{S}_{\text{global}} \leftarrow 
\cos(\mathbf{v}_{\text{cls}}, \mathbf{t}_{\text{eot}})$\;

$S_{i,j} = \cos(\mathbf{t}_i, \mathbf{v}_j)$\;
\For{$i=1$ to $M$}{
    $\mathcal{P}_i \leftarrow \arg\max_{j} \text{top-}K \, S_{i,j}$\;
    $\phi_i \leftarrow \frac{1}{K} \sum_{j \in \mathcal{P}_i} S_{i,j}$\;
}
Original local similarity $\mathcal{S}_{\text{base}} \leftarrow \frac{1}{M} \sum_{i=1}^M \phi_i$\;

\vspace{0.5em}
\textbf{Step 2: Semantic Refinement}

Extract attribute-object pairs $\mathcal{D}(T) = \{(a_m, k_m)\}$\;
\For{each $(a, k) \in \mathcal{D}(T)$}{
    $a^+ \leftarrow a$, $a^- \leftarrow \{a' | (a', k') \in \mathcal{D}(T), k' \neq k\}$\;
    $\mathbf{b}_k^+ \leftarrow \frac{1}{P} \sum_{i=1}^P [\mathcal{F}(a^+, \mathrm{obj}_i) - \mathcal{F}(\emptyset, \mathrm{obj}_i)]$\;
    $\mathbf{b}_k^- \leftarrow \frac{1}{P} \sum_{i=1}^P [\mathcal{F}(a^-, \mathrm{obj}_i) - \mathcal{F}(\emptyset, \mathrm{obj}_i)]$\;
    $\mathbf{t}'_k \leftarrow \mathbf{t}_k + \mathbf{b}_k^+ - \mathbf{b}_k^-$\;
    $\mathbf{t}'_a \leftarrow \mathbf{t}_a + \mathbf{t}_k$\;
}
\vspace{0.5em}
\textbf{Step 3: Refined Local Similarity}

Compute $\mathbf{S}' \in \mathbb{R}^{M \times N}$, $S'_{i,j} = \cos(\mathbf{t}'_i, \mathbf{v}_j)$\;
\For{$i=1$ to $M$}{
    $\mathcal{P}_i^{\mathcal{R}} \leftarrow \arg\max_{j} \text{top-}K \, S'_{i,j}$\;
    $\phi_i^{\mathcal{R}} \leftarrow \frac{1}{K} \sum_{j \in \mathcal{P}_i^{\mathcal{R}}} S'_{i,j}$\;
}
Refined local similarity $\mathcal{S}_{\text{Refine}} \leftarrow \frac{1}{M} \sum_{i=1}^M \phi_i^{\mathcal{R}}$\;

\vspace{0.5em}
\textbf{Step 4: Calculating Final Score}

$\Delta (I,T) \leftarrow \| \mathcal{S}_{\text{Refine}} - \mathcal{R}_{\text{base}} \| $\;
$\mathcal{S}_{\text{local}} \leftarrow \mathcal{S}_{\text{Refine}}  + \Delta (I,T)$\;
$\mathcal{S}_{\mathrm{final}} \leftarrow  (1-\omega) \cdot \mathcal{S}_{\text{local}} + \omega \cdot \mathcal{S}_{\text{global}} $\;

\Return $\mathcal{S}_{\mathrm{final}}$
\end{algorithm}

\subsection{Calculating Final score}
To assess whether the semantic refinement mechanism in Section~\ref{sec:semantic-refinement-mechanism} guides each text token toward the correct visual evidence, we evaluate local token--patch alignment \emph{before} and \emph{after} refinement.

Specifically, we define the binding diffence score by 
\begin{equation}
\Delta (I,T) \;=\; \| \mathcal{S}_{\text{refine}}(I,T)\;-\;\mathcal{S}_{\text{base}}(I,T) \|,
\end{equation}
which quantifies the absolute value of the difference between the refined local similarity score and original local similarity score. 

Building upon the token-level view, we construct a local matching score that combines the token-patch similarity after semantic refinement and the binding difference score that quantifies the effectiveness of semantic refinement. 
Concretely, we define
\begin{equation}
\label{eq:local-score}
\mathcal{S}_{\text{local}}(I,T) = \mathcal{S}_{\text{Refine}}(I,T) + \Delta (I,T),
\end{equation}

From the global view, the global similarity score is calculated as the standard CLIP calculation: 
\begin{equation}
\label{eq:global-score}
\mathcal{S}_{\mathrm{global}}(I, T)
= \frac{\mathbf{t}_{\mathrm{eot}}^{\top}\mathbf{v}_{\mathrm{cls}}}
       {\lVert \mathbf{t}_{\mathrm{eot}}\rVert_2 \,\lVert \mathbf{v}_{\mathrm{cls}}\rVert_2 },
\end{equation}

The final image-text matching score is obtained by fusing the global and local similarities:
\begin{equation}
\label{eq:final-score}
\mathcal{S}_{\mathrm{final}}(I,T)
= (1-\omega)\,\mathcal{S}_{\mathrm{local}}(I,T) + \omega\,\mathcal{S}_{\mathrm{global}}(I,T),
\end{equation}
where $\omega\in[0,1]$ is hyperparameter to balance the contributions of the global and local alignment scores. Algorithm~\cref{alg:abe-clip} summarizes the proposed method.
\section{Experiments}
\subsection{Experiment Settings}

\subsubsection{Datasets}
We evaluate the proposed method on three benchmarks for attribution understanding: ARO-A~\cite{yuksekgonul2022and}, SugarCrepe(swap\_att)~\cite{hsieh2023sugarcrepe} and ABC-6K~\cite{feng2022training}. Additionally, we conduct text-to-image retrieval (T2I) experiments on Flickr30K~\cite{young2014image} and MSCOCO~\cite{lin2014microsoft}.

\paragraph{ARO-A.}
ARO-A is the attribution split of the Attribution, Relation, and Order (ARO) benchmark~\cite{yuksekgonul2022and}, comprising 117 unique attribute pairs with 28{,}748 test cases in total. Each example involves two objects ($O_1$, $O_2$) and two attributes ($A_1$, $A_2$): the positive caption follows ``the $A_1$ $O_1$ and the $A_2$ $O_2$,'' while the negative caption swaps the attributes, yielding ``the $A_2$ $O_1$ and the $A_1$ $O_2$.''

\paragraph{SugarCrepe.}
The dataset employs large language models to generate grammatically correct and semantically plausible hard negatives by replacing, swapping, or adding object, attribute, and relation. We evaluate on the Swap-Attribute subset SugarCrepe(swap\_att), where captions are constructed by swapping attributes between objects.

\paragraph{ABC-6K (Attribute Binding Contrast Set).}
The positive captions in ABC-6K are collected from natural MSCOCO-2014~\cite{lin2014microsoft} captions that contain at least two color modifiers attached to different objects, while the negative counterparts are created by swapping the two color terms. Because ABC-6K is primarily introduced as a benchmark for text-to-image synthesis, with prompt-only annotations and no image identifiers, we mapped each prompt to its source MSCOCO image and constructed 3{,}213 image--text pairs.



\subsubsection{Compared Methods}
We compare the proposed method  with two baselines and several representative prior methods in compositional image--text matching.
Specifically, we adopt CLIP as the baseline model and evaluate ViT-B/16 and ViT-L/14 vision encoders~\cite{dosovitskiy2020image} with CLIP’s standard 12-layer Transformer text encoder.
We also use FG-CLIP as the global-matching baseline, using the official implementation of FG-CLIP and the publicly available pre-trained weights, which retain the fundamental architectural framework of CLIP's visual and textual encoders. We employ two types of vision encoders: ViT-B/16 and ViT-L/14~\cite{dosovitskiy2020image}, together with a base transformer with 63M parameters as text encoder.  In addition, we compare against several representative compositional mage--text matching methods, including NegCLIP, Structure-CLIP, CE-CLIP, LABCLIP, and DeGLA.
\subsection{Results}
\subsubsection{Attribute Binding Evaluation}
 \begin{table}[t]
  \centering
  \caption{Results(\%) of our method, baselines and representative prior methods on the ARO-A, SugarCrepe(swap\_att) and ABC-6K benchmarks. The best and second-best results are highlighted in \textbf{bold} and \underline{underlined}, respectively.}
  \label{exp:result1}
  \setlength{\tabcolsep}{5pt}
  \small
  \begin{tabular}{lccc}
    \toprule
    \textbf{Method} & \textbf{ARO-A} &\shortstack[c]{\textbf{SugarCrepe}} &\textbf{ABC-6K} \\
    \midrule
     CLIP$_{\text{B/16}}$          & 61.85 & 64.86 & 63.18 \\
     CLIP$_{\text{L/14}}$           & 61.63 & 62.46 & 62.43 \\
     FG-CLIP$_{\text{B/16}}$         & 64.69 & 71.02 & 72.27 \\
     FG-CLIP$_{\text{L/14}}$          & 64.34 & 69.82 & 71.62 \\
    \midrule
    NegCLIP(2023)              & 70.78 & 75.23 & 73.51 \\
    Structure-CLIP(2024)       & 80.5 & 80.5 & - \\
    CE-CLIP(2024)              & 76.4 & 77 & 75.23 \\
    LABCLIP(2025)              & 68.49 & 74.62 & 68.91 \\
    DeGLA(2025)                & 74.3 & \textbf{82.1} & 77.12 \\
    \midrule
    \textbf{ABE-CLIP$_{\text{B/16}}$}   & 67.12 & 64.86 & 66.26 \\
    \textbf{ABE-CLIP$_{\text{L/14}}$}   & 62.80 & 63.36 & 63.03 \\
    \textbf{ABE-FG-CLIP$_{\text{B/16}}$} & \textbf{84.49} & \underline{81.98} & \textbf{85.53} \\
    \textbf{ABE-FG-CLIP$_{\text{L/14}}$}  & \underline{83.75} & 80.78 & \underline{84.94} \\
    \bottomrule
  \end{tabular}
\end{table}

\begin{figure}[t]
  \centering
  \includegraphics[width=\linewidth]{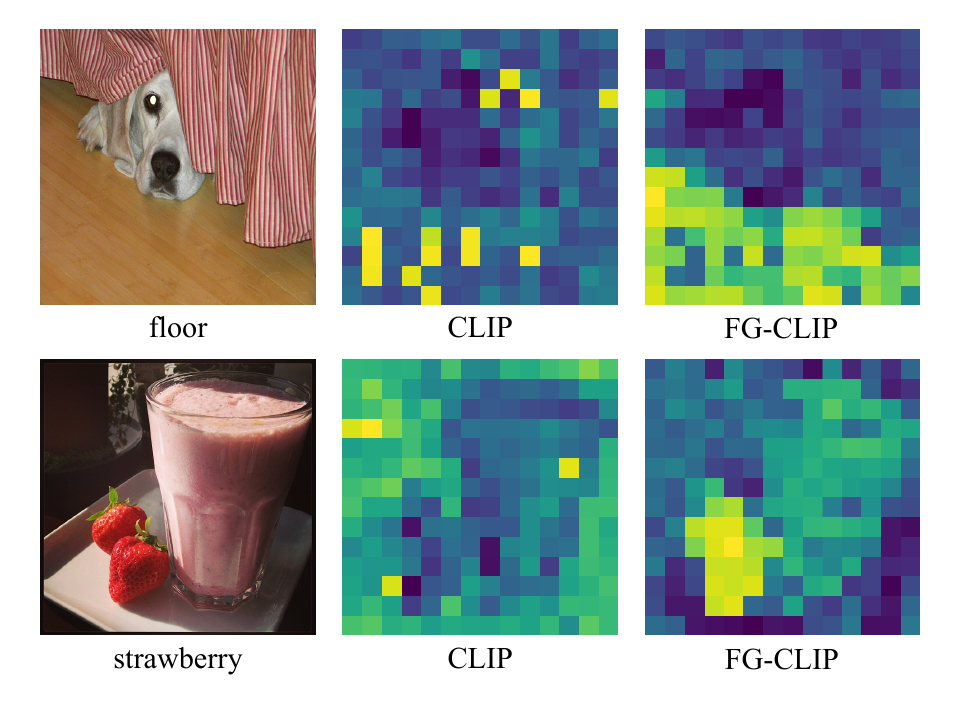}
  \caption{A comparison of similarity matrix visualizations for CLIP and FG-CLIP. We compute the similarity matrix using the words ``floor'' and ``strawberry'' with each image patch, respectively. It can be observed that CLIP fails to accurately identify the target ``floor'' and ``strawberry'', while FG-CLIP captures some relevant visual patches.}
  \label{fig:fgclip_local_alignment}
\end{figure}

\begin{table}[t]
  \centering
  \caption{Zero-shot performance(\%) on text-to-image retrieval on Flickr30K and MSCOCO (R@K). The best performance is highlighted in \textbf{bold}.}
  \label{tab:zero_shot_results}
  \setlength{\tabcolsep}{5pt}
  \scriptsize
  \begin{tabular}{@{}lrrrrrr@{}}
    \toprule
    {Method} & \multicolumn{3}{c}{Flickr30k 1K Test} & \multicolumn{3}{c}{MSCOCO 5K Test} \\
    \cmidrule(lr){2-4}\cmidrule(lr){5-7}
    & R@1 & R@5 & R@10 & R@1 & R@5 & R@10 \\
    \midrule
    CLIP$_{\text{B/16}}$            & 62.24 & 85.60 & 91.92 & 33.03 & 58.38 & 69.09 \\
    CLIP$_{\text{L/14}}$            & 66.84 & 88.98 & 93.42 & 37.04 & 61.58 & 71.56 \\
    FG-CLIP$_{\text{B/16}}$         & 74.40 & 92.12 & 95.50 & 44.36 & 69.73 & 79.55 \\
    FG-CLIP$_{\text{L/14}}$         & 81.30 & 95.52 & 97.72 & 50.46 & 74.94 & 82.96 \\
    NegCLIP                         & 65.70 & 89.46 & 93.74 & 41.51 & 68.44 & 78.62 \\
    LABCLIP                         & 63.36 & 87.66 & 92.88 & 39.78 & 67.01 & 77.46 \\
    \midrule
    \textbf{ABE-CLIP$_{\text{B/16}}$}      & 62.54 & 85.84 & 92.06 & 33.34 & 58.73 & 69.28 \\
    \textbf{ABE-CLIP$_{\text{L/14}}$}      & 67.68 & 89.38 & 93.74 & 37.26 & 61.85 & 72.00 \\
    \textbf{ABE-FG-CLIP$_{\text{B/16}}$}   & 75.22 & 92.58 & 95.78 & 45.30 & 70.72 & 80.07 \\
    \textbf{ABE-FG-CLIP$_{\text{L/14}}$}   & \textbf{81.60} & \textbf{95.62} & \textbf{97.72} & \textbf{50.78} & \textbf{75.13} & \textbf{83.11} \\
    \bottomrule
  \end{tabular}
\end{table}
We evaluate compositional attribute binding of the proposed method with standard baselines and representative prior approaches on SugarCrepe(swap\_att)~\cite{hsieh2023sugarcrepe}, ARO-A~\cite{yuksekgonul2022and} and ABC-6K~\cite{feng2022training}. We follow the evaluation protocol of prior benchmarks~\cite{hsieh2023sugarcrepe}, where a prediction is correct if the similarity score for the positive caption is higher than for the negative caption.
The results are displayed in Table~\ref{exp:result1}. We performed a grid search over the parameter $\omega$
 on the datasets and report the best setting, setting $\omega = 0.3$ in our experiments. The parameter $K$ is default to 5.
 ABE-FG-CLIP$_{\text{B/16}}$ achieves the best performance on ARO-A and ABC-6K, and performs comparably to DeGLA on SugarCrepe(swap\_att).
 Specifically, compared to its baseline FG-CLIP$_{\text{B/16}}$ , it improves 19.80\% on ARO-A, 10.96\% on SugarCrepe(swap\_att), and 13.26\% on ABC-6K. Similar gains appear on FG-CLIP$_{\text{L/14}}$  with 19.41\%, 10.96\% and 13.32\% improvements respectively. 
Notably, ABE-FG-CLIP $_{\text{B/16}}$ also substantially outperforms the methods that require extensive training, such as NegCLIP and LABCLIP. For example, it gains an improvement of 13.71\% on ARO-A, 6.75\% on SugarCrepe(swap\_att), and 12.02\% on ABC-6K compared to NegCLIP. 
When compared to vanilla CLIP$_{\text{B/16}}$, ABE-CLIP$_{\text{B/16}}$ shows only minimal improvements, a stark contrast to the remarkable gains achieved by ABE-FG-CLIP. We hypothesize this discrepancy arises because vanilla CLIP relies primarily on global image-text alignment and lacks effective fine-grained local token-patch alignment, leading our local alignment mechanism to perform suboptimally. This hypothesis is supported by the visualizations in \cref{fig:fgclip_local_alignment}, which clearly illustrate the weakness of vanilla CLIP's local alignment.
Across backbones, enlarging the architecture from ViT-B to ViT-L does not consistently improve accuracy on compositional benchmarks. This echoes prior work showing that merely increasing model capacity does not guarantee better attribute binding.

\subsubsection{Zero-Shot Cross-Modal Retrieval Evaluation}
To evaluate the performance of our approach in text-to-image retrieval, we conduct experiments on the Flickr30K and MSCOCO benchmarks.
We report the commonly used Recall@$K$ (R@$K$) on the Flickr30K-1K test split and the MSCOCO 5K test set, as shown in Table~\ref{tab:zero_shot_results}. From the table, we can see that ABE-CLIP and ABE-FG-CLIP achieves slightly better performance compared to their baselines CLIP and FG-CLIP, respectively. And ABE-FG-CLIP$_{\text{L/14}}$ achieves the best result among all compared methods. These results suggest that our method is also competitive for general image--text retrieval tasks.

\subsubsection{Ablation Study}
\begin{figure}[t]
  \centering
  \includegraphics[width=0.85\linewidth]{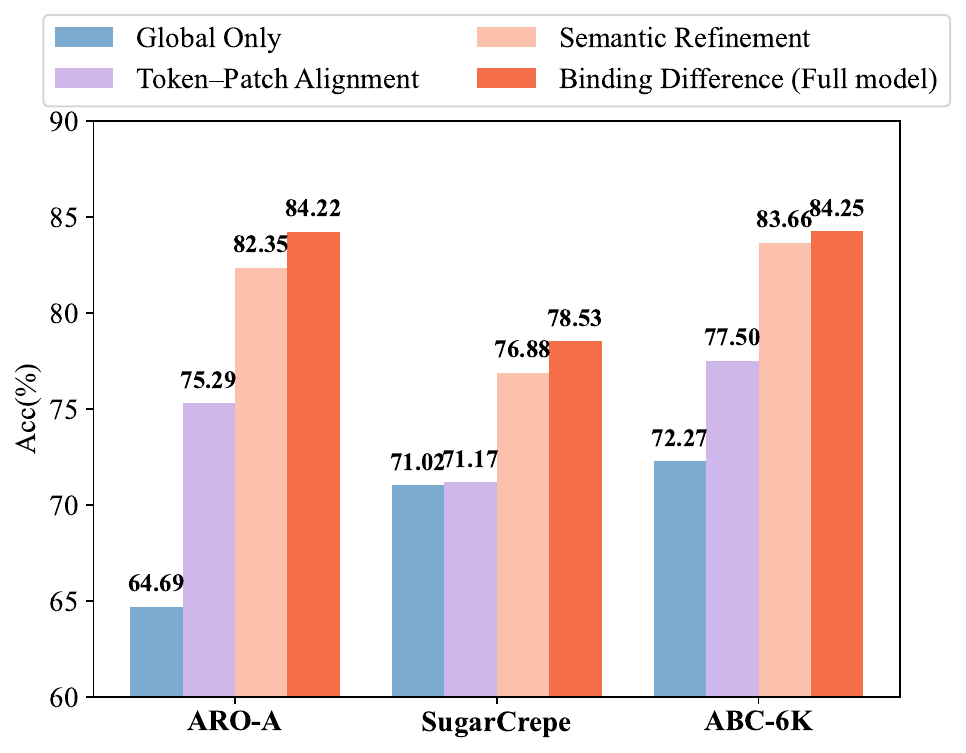}
  \caption{Results of ablation study on different components. Incorporating Local Token–Patch Alignment, the Semantic Refinement Mechanism, and Binding Difference Scores yields progressive, incremental performance gains.}
  \label{fig:bar_chart}
\end{figure}

\paragraph{Component Analysis.}
We study the contribution of each component in ABE-CLIP, i.e., Local Token--Patch Alignment, Semantic Refinement Mechanism and Binding Difference Score. Experiments are conducted on ABE-FG-CLIP\textsubscript{B/16}. As shown in \cref{fig:bar_chart}, the results show that all four components are beneficial for the three benchmarks. Compared to the baseline using global representation for similarity calculation, Local Token--Patch Alignment yields substantial gains on ARO-A (+10.60\%), and ABC-6K (+5.23\%), with a modest change on SugarCrepe(swap\_att) (+0.15\%). This indicates that global similarity alone is insufficient for capture subtle semantic difference. Introducing the Semantic Refinement Mechanism further improves performance, with an absolute gain of +7.06\% on ARO-A, +5.71\% on SugarCrepe(swap\_att), and +6.16\% on ABC-6K, respectively. This suggest that the Semantic Refinement Mechanism can effectively disentangle object-attribute relationships, enhancing the effect of local alignment of text token and image patches, and therefore improving the final results. Besides, adding the Binding Difference Score provides additional gains over the Semantic Refinement Mechanism: +1.87\% on ARO-A, +1.65\% on SugarCrepe(swap\_att), and +0.59\% on ABC-6K, respectively. 

\paragraph{Robustness analysis on top-$K$ Patch Pooling.}
For each text token, the token-wise score is computed as the the mean similarity to its top-$K$ most similar image patches. We investigate how the number of selected patches affects performance by varying $K\in\{1,3,5,8,10\}$. 
As shown in Table~\ref{tab:patch_nums_effect}, increasing $K$ from 1 to 5 consistently improves performance: +0.93\% on ARO-A, +0.15\% on SugarCrepe(swap\_att), and +0.81\% on ABC-6K. Further increasing $K$ to 10 results in performance saturation or slightly decline: $-0.17\%$ (ARO-A), $-0.90\%$ (SugarCrepe(swap\_att)), and $-0.25\%$ (ABC-6K). This indicates that smaller $K$ restricts the pool of spatial cues available for alignment, while larger $K$ admits more background or distractor patches that dilute token--patch alignment. 
The overall effect of $K$ on the results is not significant when $K$ is larger than $5$, which indicates that our method has strong robustness to the parameter $K$.  

\begin{table}[t]
  \centering
  \caption{Robustness of our method on top-$K$ patch pooling. This table shows the results (\%) on ARO-A, SugarCrepe (swap\_att), and ABC-6K for different values of $K$ from 1 to 10. The best results are highlighted in \textbf{bold}.}
  \label{tab:patch_nums_effect}
  \small
  \setlength{\tabcolsep}{4pt} 
  \begin{tabular}{@{}lccc@{}}
    \toprule
    \textbf{$K$} & \textbf{ARO-A} & \textbf{SugarCrepe (swap\_att)} & \textbf{ABC-6K} \\
    \midrule
    1  & 83.56 & 81.83 & 84.72 \\
    3  & 84.33 & 81.98 & 85.47 \\
    5  & \textbf{84.49} & \textbf{81.98} & \textbf{85.53} \\
    8  & 84.47 & 81.53 & 85.47 \\
    10 & 84.32 & 81.08 & 85.28 \\
    \bottomrule
  \end{tabular}
\end{table}


\section{Conclusion}
In this paper, we present \textbf{ABE-CLIP}, a training-free, plug-and-play method aimed at enhancing attribute--object binding in compositional image--text matching. We introduce Local Token--Patch Alignment which refines the scoring by computing token-level scores via top-$K$ patch aggregation, yielding finer-grained alignment and composition-aware similarity. To obtain finer-grained concept disentanglement, we propose Semantic Refinement Mechanism which modify the embeddings of objects and attributes.

While our work specifically addresses the attribute-binding problem in complex multi-object scenes, it does not explore other key aspects of compositional understanding, such as spatial reasoning and relational reasoning. Notably, our method holds significant potential for being extended to enhance other compositional understanding capabilities. In future work, we will explore leveraging this potential to extend our method to these capabilities, aiming to tackle a broader spectrum of visual tasks.
